\documentclass{article}

\PassOptionsToPackage{numbers, compress}{natbib}
\usepackage[preprint]{neurips_2025}

\usepackage[utf8]{inputenc} 
\usepackage[T1]{fontenc}    
\usepackage{hyperref}       
\usepackage{url}            
\usepackage{booktabs}       
\usepackage{amsfonts}       
\usepackage{nicefrac}       
\usepackage{microtype}      
\usepackage{xcolor}         
\usepackage{tabularx}
\usepackage[T1]{fontenc}
\usepackage{graphicx} 
\usepackage{multirow}
\usepackage{array}
\usepackage{booktabs}
\usepackage{adjustbox}
\usepackage{graphicx}
\usepackage{subcaption}
\usepackage{epigraph}
\usepackage{tcolorbox}
\definecolor{RoyalBlue}{RGB}{65,105,225}
\definecolor{Green}{RGB}{34,139,34}
\hypersetup{
    colorlinks=true,
    citecolor=RoyalBlue,
    linkcolor=RoyalBlue,
    filecolor=magenta,      
    urlcolor=RoyalBlue
}
\usepackage{multicol}
\usepackage{graphicx,wrapfig}
\usepackage{enumitem}
\usepackage{amsmath,amsbsy,amsfonts,amssymb,amsthm}
\usepackage{bm}
\usepackage{newtx}
\usepackage{algorithm,algorithmic,mathtools}
\usepackage{color,cases}
\usepackage{tikz,bbm}
\usepackage[nameinlink]{cleveref}
\usetikzlibrary{calc,shapes}
\usepackage{hyperref,comment}
\usepackage{xcolor,url,verbatim}
\usepackage{afterpage}
\usepackage{placeins}

\crefname{claim}{Claim}{Claims}

\crefname{lemma}{Lemma}{Lemma}

\crefname{theorem}{Theorem}{Theorem}

\newtheorem{remark}{Remark}

\title{MEMETRON: Metaheuristic Mechanisms for Test-time Response Optimization of Large Language Models}

\author{
  Son The Nguyen \\
  Department of Information and Decision Sciences\\
  University of Illinois Chicago\\
  \texttt{snguye65@uic.edu}
  \And
  Theja Tulabandhula \\
  Department of Information and Decision Sciences\\
  University of Illinois Chicago\\
  \texttt{theja@uic.edu}
}

\begin{document}

\maketitle

\begin{abstract}

Large language models (LLMs) are increasingly used for both open-ended and structured tasks, yet their inference-time behavior is still largely dictated by heuristic decoding strategies such as greedy search, sampling, or reranking. These methods provide limited control and do not explicitly optimize for task-specific objectives. We introduce MEMETRON, a task-agnostic framework that formulates LLM decoding as a discrete black-box optimization problem. MEMETRON leverages hybrid metaheuristic algorithms, GENETRON and ANNETRON, to search the response space, guided by reward models and contextual operations performed by the LLM itself. This approach enables efficient discovery of high-reward responses without requiring model retraining or gradient access. The framework is modular and generalizes across diverse tasks, requiring only a reward function and lightweight prompt templates. We evaluate our framework on the critical human preference alignment task and demonstrate that it significantly outperforms standard decoding and reranking methods, highlighting its potential to improve alignment without model retraining.

\end{abstract}

\begin{sloppypar}
\section{Introduction}

Large language models (LLMs) have become central to many NLP and real-world applications, from open-ended dialogue and summarization to agentic systems that perform autonomous tasks such as decision-making, code generation, and workflow automation. Enabling these capabilities depends on effective alignment methods, yet current approaches have seen limited deployment beyond improving broad traits such as helpfulness, safety, and reasoning. Train-time methods such as Reinforcement Learning from Human Feedback (RLHF) \citep{ouyang2022training,christiano2017deep}, the direct finetuning variant Direct Preference Optimization (DPO) \citep{rafailov2023direct}, and the more recent Group Relative Policy Optimization (GRPO) \citep{shao2024deepseekmath} are capable of optimizing more complex behaviors, but they are technically demanding, time-consuming, and computationally and resource-intensive to train. These methods typically rely on data sampled from base models and ranked by humans or learned reward models. Because this data is collected passively rather than actively selected to maximize a reward function, it may not include the most effective outputs for the target objective. Furthermore, since these methods require offline optimization and access to model gradients, they are typically applied by experts in technically demanding training environments. This limits their flexibility and adaptability to end user-specific goals at inference/test-time. While some recent efforts explore iterative or online variants of RLHF and DPO to address this limitation \citep{zhang2024self,dong2024rlhfworkflowrewardmodeling,pang2024iterativereasoningpreferenceoptimization}, these approaches are even more resource-intensive and remain uncommon in practice.

At test-time, users often rely on heuristic decoding strategies, such as greedy or beam search and temperature‑ or top‑\(k\)/top‑\(p\) sampling, to produce a single response. Simple enhancements, such as best‑of‑\(n\) sampling, self‑consistency \citep{wang2023selfconsistencyimproveschainthought}, prompt ensembling \citep{jiang2023llmblender}, or self‑refinement \citep{madaan2023selfrefineiterativerefinementselffeedback}, can yield incremental gains but remain one‑shot or shallow: they generate a fixed set of candidates and select or merge without an iterative, reward‑guided feedback loop. Even Mixture‑of‑Agents (MoA) \citep{wang2024moa}, which iterates through multiple layers, lacks an explicit objective function and stopping criterion. Consequently, these methods cannot guarantee progress toward maximizing an arbitrary, black‑box reward.

Recent work on search-based methods with reranking, such as beam search and lookahead search with verifiers, offer a more principled alternative \citep{wang2023math, lightman2023let, snell2024scalingllmtesttimecompute}. These approaches explore a broader candidate space and incorporate feedback during generation, enabling more structured optimization. They are particularly effective in domains like mathematics and multi-step reasoning, where verifying or evaluating intermediate steps, often structured as trees, can guide the model toward more reliable and accurate final solutions.

We complement this line of work by treating inference-time LLM sampling as a discrete search problem over full candidate responses. This makes it applicable across a broader range of tasks, including open-ended generation, where intermediate verification is not always feasible. Rather than tuning model weights or relying solely on probability-based heuristics, we design our algorithms based on metaheuristic optimization strategies, augmented by LLM's generative and semantic strengths. Our framework, MEMETRON, combines global exploration and local refinement.
\begin{itemize}
  \item \textbf{GENETRON} (Global Exploration) evolves a population of candidate responses using evolutionary operators, with the LLM acting as a semantic crossover operator.
  \item \textbf{ANNETRON} (Adaptive Local Search) refines individual responses via simulated annealing–style moves, using the LLM as a context‑aware annealing operator.
  \item \textbf{MEMETRON} alternates between GENETRON's global steps and ANNETRON's local refinements in a unified loop.
\end{itemize}

\begin{figure}
    \centering
    \includegraphics[width=1\linewidth]{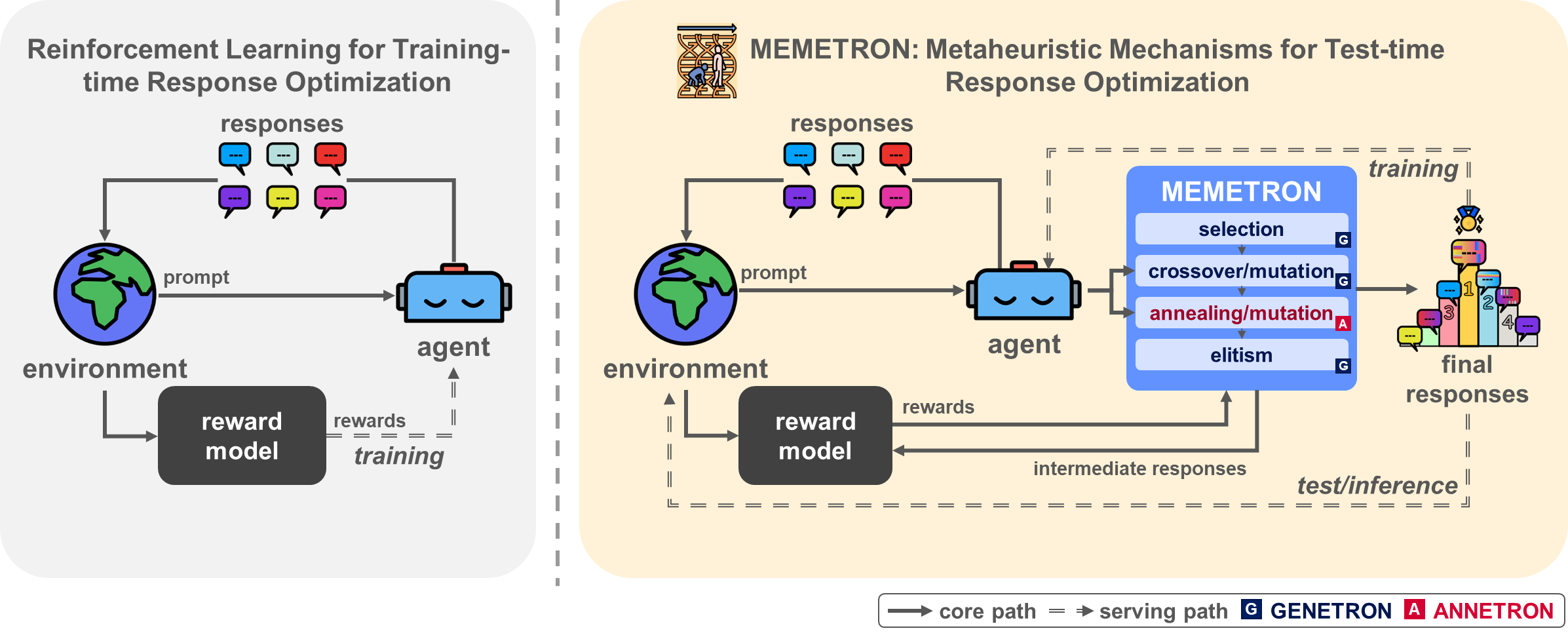}
    \caption{MEMETRON, composed of GENETRON and ANNETRON, iteratively optimizes LLM responses at test-time using reward model(s), without accessing the model’s internals. The optimized responses and history buffer can be used either to serve end users (i.e., the environment) or to further improve models during training via methods such as SFT, RLHF, DPO, or GRPO.}
    \label{fig:enter-label}
\end{figure}
\noindent \textbf{Our main contributions are:}
\begin{enumerate}
  \item We formally define test-time LLM decoding as a discrete black-box optimization problem:
  \[
    \max_{y \in \mathcal{Y}}\,r(x,y),
  \]
  where \(\mathcal{Y}\) is the discrete set of sequences producible by any decoding strategy, and \(r(x, y)\) is a user-defined reward function.

  \item We introduce \textbf{MEMETRON}, a task-agnostic hybrid framework for optimizing LLM outputs at inference time. MEMETRON consists of two core algorithms, \textbf{GENETRON} and \textbf{ANNETRON}, which perform search over \(\mathcal{Y}\). The framework is carefully designed to invoke the LLM only for context-sensitive or coherence-critical operations, while incorporating classical or sampling-based operators where appropriate.
  
  \item The proposed algorithms can be applied ``as‑is'' individually or together to any discrete LLM‑response optimization task: users simply supply the task‑specific reward function \(r(x,y)\) and minimal prompt templates. 

  \item We enable exploration during inference: GENETRON maintains evolving populations, while ANNETRON performs stochastic local moves. This allows the discovery of high-reward outputs that are typically missed by one-shot decoding or reranking.

  \item This approach not only serves as an effective method for test-time optimization, but also supports integration into training pipelines such as supervised fine-tuning (SFT), RLHF, DPO, and GRPO.

  \item We validate MEMETRON on tasks such as human preference alignment to validate our claims.

\end{enumerate}

Our objective is not to find the global optimum in the response space, which is generally intractable for high-dimensional language outputs. Instead, MEMETRON aims to discover responses that are meaningfully better than those produced by standard one-shot decoding or shallow reranking methods, given a fixed budget of model calls and reward evaluations.

The remainder of the paper is organized as follows. Section \ref{relatedwork} reviews related work on LLM decoding, test‑time search methods, and metaheuristic techniques. Section \ref{problemsetup:3.1notation} presents our formal Problem Formulation and section \ref{algo} details the algorithms. Section \ref{experiments} describes experimental setups and results. Finally, Section \ref{conclusions} concludes with discussions and future directions.

\section{Related Work} \label{relatedwork}

\subsection{Training-Time Response Optimization}
Methods such as Reinforcement Learning from Human Feedback (RLHF) \citep{ouyang2022training, christiano2017deep} and Direct Preference Optimization (DPO) \citep{rafailov2023direct} optimize for preferred outputs by updating model parameters using gradient-based techniques. In RLHF framework, a reward model trained on human preference data is used to guide policy optimization via reinforcement learning algorithms such as Proximal Policy Optimization (PPO) \citep{schulman2017proximal}, where gradients are computed with respect to the model's parameters. DPO, by contrast, avoids explicit reward modeling. It formulates the preference task as a classification-like loss over pairs of preferred and dispreferred responses, enabling direct supervised fine-tuning via standard gradient descent. Despite differences in formulation, both of these approaches, among other of their variations, require access to model internals and retraining to steer language model behavior toward more desirable outputs. 

Due to the computational cost, these optimization steps are typically performed offline using pre-collected datasets, often with only a limited number of fine-tuning epochs. As a result, RLHF and DPO mainly operate in an exploitation regime, with limited or no exploration during training, though some iterative approaches do exist \citep{zhang2024self,dong2024rlhfworkflowrewardmodeling,pang2024iterativereasoningpreferenceoptimization}.

\subsection*{RL–Induced Inference Behavior}
While RL is often used to align LLMs with human preferences or task-specific behaviors during training, it has also been applied to encourage structured reasoning capabilities. In this setting, the objective is to shape internal policies that generalize to more deliberative inference-time behavior. For example, OpenAI's \texttt{o1} \citep{openai2024o1systemcard} and \texttt{o3} \citep{openai2024o3} models are trained using RL to internalize behaviors such as producing intermediate reasoning steps before arriving at a final response. Recently, DeepSeek's R1 model employs Group Relative Policy Optimization (GRPO) \citep{shao2024deepseekmath}, to enhance its ability to perform complex reasoning tasks. Although no explicit reasoning instructions are given at inference time, the learned policies lead the model to exhibit structured, multi-step reasoning test-time behaviors. 

This approach requires extensive training with significant computational resources, making them costly to develop and scale.

\subsection{Test-Time Response Optimization}
Test-time compute techniques have emerged as a complementary strategy to training-time methods \citep{snell2024scalingllmtesttimecompute}. These approaches leverage additional computation during inference to improve model outputs without modifying the underlying weights. They are particularly useful when retraining is impractical, making them applicable across various deployment settings.

\textbf{Prompt Engineering.}
Prompting is one of the earliest and most widely used forms of test-time optimization. It guides the model to produce better outputs through carefully designed input formats. Chain-of-Thought (CoT) prompting encourages step-by-step reasoning through structured examples \citep{wei2022cot}. Self-refinement prompts the model to iteratively revise its own outputs, often improving quality without external feedback \citep{madaan2023selfrefineiterativerefinementselffeedback}. Least-to-most prompting decomposes complex tasks into simpler sub-questions solved sequentially \citep{zhou2023leasttomostpromptingenablescomplex}. 

However, prompt-based strategies typically lack a clear success signal at inference time, making their effectiveness hard to verify. While many prompting techniques are general and lightweight, they do not incorporate explicit optimization toward a reward function. As a result, they are often evaluated qualitatively or through task-specific studies in controlled settings. 

\textbf{Search-Based Methods.}
Search-based techniques generate multiple candidate responses and then select the most promising one based on some scoring criterion. This score-and-select approach can use heuristics, log probabilities, verifiers, or learned reward models to evaluate candidates. Examples include best-of-n sampling \citep{ouyang2022training} and self-consistency \citep{wang2023selfconsistencyimproveschainthought}, beam search, lookahead search with verifiers \citep{snell2024scalingllmtesttimecompute}, or MCTS. 

\textbf{Ensembling Techniques.}
Ensembling methods aim to combine the strengths of multiple models or responses. A common strategy is score-and-merge: for example, LLM-Blender \citep{jiang2023llm} uses a pairwise ranking model to identify top responses, then merges them using a generative fusion model. While effective, this approach may discard useful lower-ranked responses, limiting diversity.

Simple methods like best-of-n sampling, self-consistency, or LLM-Blender are typically one-shot and shallow. They generate a fixed set of outputs and rerank or aggregate them without iterative feedback. In contrast, more structured methods like beam search, lookahead search with verifiers, and MCTS perform step-wise exploration over reasoning trees or action sequences, allowing more deliberate optimization during generation.

The Mixture-of-Agents (MoA) approach \citep{wang2024moa} falls under both ensembling and prompt engineering. It addresses the limitations of relying solely on top-ranked outputs, as in \citep{jiang2023llm}, by generating responses from multiple LLMs and synthesizing them into final outputs through one or more aggregator LLMs, often across multiple layers. Unlike score-and-merge strategies, MoA does not rank or filter candidates; instead, it leverages the aggregators' generative capabilities to directly integrate the full set of responses. While this allows for the incorporation of a broader range of perspectives and reasoning paths, it also introduces several challenges, most notably, context length limitations due to the inclusion of many intermediate responses, the lack of an explicit optimization objective to guide synthesis, and the absence of a principled stopping criterion.

In contrast to prior work, our approach introduces a general-purpose, reward-guided search framework that operates directly over the space of full model outputs. Unlike train-time alignment methods such as RLHF and DPO, it requires no gradient access or retraining. Compared to prompting and shallow search techniques, it offers explicit optimization toward user-defined objectives. And unlike step-wise tree search or agent-based ensembling, our method performs structured exploration over complete outputs, making it broadly applicable to different tasks. Furthermore, our approach ensures that a diverse set of candidate responses across the ranking are fused and refined, without incurring the context-length overhead of approaches like MoA. Critically, this refinement is not heuristic or static, but is continuously guided by reward models throughout the search process.

\textbf{Metaheuristic Search Strategies.} Genetic Algorithms (GAs) \citep{holland1975adaptation} are population-based metaheuristics inspired by natural selection, employing operators such as selection, crossover, and mutation to evolve candidate solutions. Simulated Annealing (SA) \citep{kirkpatrick1983optimization}, in contrast, is a single-solution stochastic method that mimics the physical annealing process to escape local optima by probabilistically accepting worse solutions. While both approaches are effective for combinatorial and continuous optimization, they exhibit different strengths: GAs explore broadly through population diversity, whereas SA emphasizes local refinement with a controlled exploration-exploitation trade-off. Memetic Algorithms (MAs) \citep{moscato1989evolution}, also known as hybrid GAs, combine the global search capabilities of GAs with local search techniques, often incorporating domain-specific heuristics, to achieve more efficient convergence. MAs have shown improved performance across various problem domains by balancing exploration and exploitation more effectively than either GA or SA alone \citep{moscato1989evolution,5447938,1554795}.

Metaheuristic algorithms have also been explored in recent studies. For Genetic Algorithms (GA), \citep{hemberg2024evolvingcodelargelanguage} leverages LLMs as all operators to evolve code solutions. Similarly, \citep{lee2025evolvingdeeperllmthinking} proposed an evolutionary search strategy for planning tasks. However, these methods are adapted to very specific cases where ground truths are available to compare with LLM outputs. For Simulated Annealing (SA), \citep{xu2024llmrefinepinpointingrefininglarge} trains a fine-grained feedback model to provide precise information about problems in the generated output. This feedback is then converted into scalar scores using a manual rule-based scoring scheme to guide the SA process.

Our approach differs in that it does not require ground truths or additional training. We formalize and generalize both GA and SA strategies to ensure compatibility with a wide range of objectives and arbitrary reward signals, including those derived from human preference. This design follows the core characteristics of metaheuristics as high-level optimization frameworks that can be applied across different problem domains. Building on this formulation, we introduce three hybrid algorithms: \textbf{GENETRON}, \textbf{ANNETRON}, and \textbf{MEMETRON}, each designed to perform reward-guided search over LLM outputs at test-time. The framework is carefully designed to engage the LLM selectively when semantic judgment or coherence is needed, while delegating other steps to classical or sampling-based procedures, allowing for efficient search with minimal computational overhead.

\section{Problem Formulation}\label{problemsetup:3.1notation}

We frame the task of sampling responses from LLMs as a discrete optimization problem. The objective is to identify a high‑quality response according to a given reward function. Formally, given a prompt \(x\), a language model \(\pi(y \mid x)\), and a black‑box reward function 
\[
r: \mathcal{X} \times \mathcal{Y} \to \mathbb{R},
\]
which assigns a scalar reward to each response \(y \in \mathcal{Y}\), the goal is to find
\[
y^* \;=\; \arg\max_{y \in \mathcal{Y}}\,r(x, y),
\]
where the response space \(\mathcal{Y}\) is defined by all sequences producible by decoding \(\{\pi_m(y \mid x)\}_{m=1}^M\) via temperature‑controlled sampling, nucleus (top‑\(p\)) sampling, top‑\(k\) sampling, minimum‑\(p\) sampling, or beam search.

This setup presents several challenges:
\begin{itemize}
  \item \(\mathcal{Y}\) is discrete and vast (on the order of \(|V|^L\) for vocabulary size \(|V|\) and max length \(L\)),
  \item \(r(x,y)\) is a black‑box reward with no gradient information, and
  \item The objective landscape is non‑convex, non‑smooth, and highly multi‑modal (small changes in \(y\) can cause abrupt jumps in \(r\), and distinct “high‑reward” clusters may be disconnected).
\end{itemize}

Because standard optimization methods (gradient descent, convex solvers, exact combinatorial searches) are ill‑suited to this discrete, gradient‑less, multi‑modal landscape, we turn to metaheuristic algorithms, such as genetic algorithms (GA) and simulated annealing (SA), as a more practical and effective framework for approximating
\(
\arg\max_{y \in \mathcal{Y}}\,r(x, y)
\)
in large, discrete, nonconvex spaces. By leveraging stochastic sampling and population‑based heuristics, these methods explore complex search landscapes. Although metaheuristics do not guarantee a global optimum under a finite budget, mechanisms like GA’s population‑based variation or SA’s temperature‑controlled random walks enable escapes from local optima. In principle, with sufficient iterations they converge to a global optimum, but in practice we truncate under a fixed evaluation budget. 

Our goal is not to exhaustively search the entire output space or guarantee global optimality. Instead, we aim to discover higher-reward responses than those typically found by standard decoding or shallow reranking methods, while remaining computationally efficient and broadly adaptable.

\subsection{Reward Signal Design}

MEMETRON requires a user-defined reward function \( r(x, y) \) to guide search over candidate responses. This function can be flexibly defined to reflect task-specific goals or deployment constraints. Common reward signals include automatic metrics such as BLEU or ROUGE, factuality and consistency scores, toxicity penalties, domain-specific rules, and human feedback such as thumbs-up ratings or preference labels. Internal model signals like log probability can also be used, either as standalone objectives or as part of composite rewards. For example, we can define
\[
r(x, y) = \alpha \cdot r_{\text{task}}(x, y) + (1 - \alpha) \cdot \log p(y \mid x),
\]
to balance task performance with model confidence.

We assume users can define or learn task-appropriate reward functions to guide response optimization. Although reward model design is beyond the scope of this work, it remains a nontrivial challenge and has a significant impact on the effectiveness of test-time optimization. Prior work has investigated what makes a reward model an effective learning signal \citep{razin2025makes}, how overoptimization can emerge during training \citep{gao2023scaling}, and how it can be mitigated through constrained objectives or demonstration-guided approaches \citep{rita2024countering, moskovitz2023confronting}.

\section{Metaheuristics for Test-time Tesponse Optimization}\label{algo}

\subsection{GENETRON (Genetic Mechanism for Response Optimization)} \label{genetron}

We propose GENETRON,  an evolutionary algorithm, that optimizes model-generated responses by iteratively selecting, combining, and mutating a population of candidates toward higher reward. The process begins by initializing a population \( G^{(0)} = \{y_1^{(0)}, \dots, y_N^{(0)}\} \subset \mathcal{Y} \) of \( N \) candidate responses, sampled from one or multiple models \( \{\pi_m(y \mid x)\}_{m=1}^{M} \), with \( M \) denoting the total number of models and \( x \) being the input prompt. Each response \( y_i^{(0)} \in G^{(0)} \) is evaluated using a reward function \( r(x, y) \), yielding a corresponding set of scores \( R^{(0)} = \{r(x, y_1^{(0)}), \dots, r(x, y_N^{(0)})\} \). 

At each subsequent generation \( g \), a parent set \( P^{(g)} \subseteq G^{(g)} \) is selected via binary tournament selection: for each parent, two candidates are sampled randomly from the current population, and the one with the higher reward \( r(x, y) \) is retained. To generate offspring, we apply \emph{\textbf{crossover}} via \emph{\textbf{LLM-as-Crossover-Operator}}. For each offspring, a pair of parents \( y_i, y_j \in P^{(g)} \) is sampled at random, and a fusion prompt is constructed:
\begin{quote}
$x_{\text{fuse}}(x, y_i, y_j)$ = ``Task:
\begin{enumerate}
    \item You are an expert at fusing responses into a single, original response. For the query below, you have been provided with two responses from different AI models. Your goal is to learn from these responses to produce a fused response that is better than either original response.
    \item First, critically analyze both responses, including components within them, for factual accuracy, logical consistency, adherence to instructions, writing quality, potential improvements, and any biases or unfounded assumptions.
    \item Based on your analysis, identify whether the responses, or their components, conflict or complement each other. If conflicts cannot be reasonably reconciled, select the most accurate and relevant content. If the ideas are accurate and complement each other in answering the query, fuse the compatible ideas into a cohesive, well-structured response. Avoid combining contradictory points.
    \item Ensure the final response is clear, concise, well-structured, and directly addresses the query. It should be logically coherent, contextually relevant, and easy to read. Eliminate all forms of redundancy, such as repetitive information, unnecessary restatements, or duplicated content.
\end{enumerate}
Query: {x}\newline
Response 1: $y_i$\newline
Response 2: $y_j$"
\end{quote}

Offspring can then be sampled from the model via \( \pi(y \mid x_{\text{fuse}}(x, y_i, y_j)) \). To further promote diversity and encourage higher-reward candidates, \emph{\textbf{mutation}} is implemented via best-of-n sampling: from each fusion prompt, \( n \) samples are drawn,
\[
\{y_1', \dots, y_n'\} \sim \pi(y \mid x_{\text{fuse}}(x, y_i, y_j)),
\]
and the response with the highest reward is selected:
\[
y^* = \underset{{y' \in \{y_1', \dots, y_n'\}} }{\arg\max}r(x, y').
\]
The resulting set of all generated offsprings of generation \(g\) is denoted as \( O^{(g)} \). Throughout the process, we maintain a cumulative search history
\[
\mathcal{H} = G^{(0)} \cup \bigcup_{g=1}^{L} O^{(g)},
\]
where \(L\) stands for the last generation index. The next population \( G^{(g+1)} \) is formed with \emph{\textbf{elitism}}, by selecting the top \( N \) responses from the full history \( \mathcal{H}^{(g)} \), based on reward:
\[
G^{(g+1)} = \underset{\substack{S \subseteq \mathcal{H}^{(g)} \\ |S| = N}}{\arg\max} \sum_{y \in S} r(x, y)
\]
This step allows high-reward individuals from any previous generation to persist if they outperform newly generated candidates. The evolutionary process continues until a convergence criterion is met such as such as detecting diversity collapse (e.g., low variance among candidates), hitting a fixed computational budget (e.g., total number of model calls), reaching a predefined reward threshold, or a fixed number of generations \( L \) is reached. In practice, convergence is determined by monitoring the improvement in maximum reward across generations. Specifically, if the best reward in the current population \( G^{(g)} \) has not improved by more than a small threshold \( \delta \) over the past \( L_{\text{patience}} \) generations, the search is considered to have converged. Formally, we stop if
\[
\max_{y \in G^{(g)}} r(x, y) - \max_{y \in G^{(g - L_{\text{patience}})}} r(x, y) < \delta.
\]
This patience-based criterion prevents unnecessary computation once the search plateaus and further improvements are unlikely.

GENETRON's output supports a wide range of downstream \textit{\textbf{training}} objectives, including reward model tuning, preference-based learning (e.g., DPO, GRPO), and RLHF. For \textbf{\textit{test-time}} or SFT use, the highest quality response for a given prompt can be extracted from the history by computing
\[
\hat{y} = \arg\max_{y \in \mathcal{H}} r(x, y).
\]

\begin{algorithm}[htbp]
\caption{GENETRON}
\label{alg:ero}
\begin{algorithmic}[1]
\STATE $g \gets 0$
\STATE Initialize population $G^{(0)} = \{y_1^{(0)}, \dots, y_N^{(0)}\}$ by sampling from model $\pi(y \mid x)$
\STATE Evaluate rewards: $R^{(0)} = \{r(x, y_1^{(0)}), \dots, r(x, y_N^{(0)})\}$
\STATE Initialize history buffer: $\mathcal{H} \gets G^{(0)}$
\WHILE{termination criterion not met}
    \STATE \emph{Selection:} Select parents $P^{(g)} \subseteq G^{(g)}$ via binary tournament selection
    \STATE Initialize offspring set: $O^{(g)} \gets \emptyset$
    \FOR{$k = 1$ to $N$}
        \STATE \emph{Crossover:} Sample parent pair $y_i, y_j \in P^{(g)}$
        \STATE Construct fusion prompt $x_{\text{fuse}} \gets x_{\text{fuse}}(x, y_i, y_j)$
        \STATE \emph{Mutation:} Sample $n$ candidates $\{y_1', \dots, y_n'\} \sim \pi(y \mid x_{\text{fuse}})$
        \STATE Select best candidate $y^* = \underset{y' \in \{y_1', \dots, y_n'\}}{\arg\max} r(x, y')$
        \STATE Add $y^*$ to $O^{(g)}$
    \ENDFOR
    \STATE Update history: $\mathcal{H} \gets \mathcal{H} \cup O^{(g)}$
    \STATE Form next population:
        \[
        G^{(g+1)} = \underset{\substack{S \subseteq \mathcal{H}^{(g)} \\ |S| = N}}{\arg\max} \sum_{y \in S} r(x, y)
        \]
    \STATE $g \gets g + 1$
\ENDWHILE
\STATE \textbf{Output:} $\mathcal{H}$ and best response $\hat{y} = \arg\max_{y \in \mathcal{H}} r(x, y)$
\end{algorithmic}
\end{algorithm}

\begin{remark}[Analysis of GENETRON Optimization Dynamics]
GENETRON is designed to evolve high-reward responses over successive generations by structuring the search process through reward-guided genetic operations. Tournament selection biases parent choice toward higher-reward responses, increasing the likelihood that strong candidates influence the next generation. At the same time, it maintains diversity by allowing lower-ranked candidates to regularly participate, enabling the \emph{\textbf{LLM-as-Crossover-Operator}} to later leverage niche strengths or complementary features that may not be present in top performers.

The \emph{\textbf{LLM-as-Crossover-Operator}} then leverages the contextual understanding of the LLM to combine strengths and reconcile conflicts between selected parents, incentivizing more informative and higher-quality responses. Implicit mutation, implemented through best-of-\(n\) sampling, further explores the response space and looks for better or fitter responses in the local neighborhood.

Finally, elitism preserves top-performing candidates to ensure that the valuable and high-reward traits responsible for their success remain accessible to future generations. If those traits have not yet been effectively propagated or combined through crossover and mutation, elitism gives the algorithm continued opportunities to do so in the long term.
\end{remark}

\subsection{ANNETRON (Annealing Mechanism for Response Optimization)}\label{annetron}

We propose \textbf{ANNETRON}, a response optimization algorithm based on \textit{Simulated Annealing (SA)}, which performs stochastic local search over the response space of LLMs. ANNETRON enables effective exploration under black-box reward functions by balancing random exploration with reward-guided exploitation.

The optimization process begins by initializing a candidate response \( y_0 \in \mathcal{Y} \), sampled from a conditional language model \( \pi(y \mid x) \), where \( x \) is the input prompt. The corresponding reward is computed as \( r_0 = r(x, y_0) \). A positive temperature parameter \( T_0 > 0 \) is also initialized, and the search proceeds over discrete time steps indexed by \( t \).

At each iteration \( t \), a neighboring candidate \( y' \) is generated using an \emph{\textbf{LLM-as-Refinement-Operator}} \( \pi(y \mid x_{\text{refine}}(x, y_t)) \), where refinement prompt is constructed as:

\begin{quote}
$x_{\text{refine}}(x,y_t)$ = ``Task:
\begin{enumerate}
    \item You are a professional writing assistant specialized in refining responses. For the query below, you have been provided with a single response from another AI model. Your task is to compete with this response to prove that you are a more capable AI model.
    \item Critically analyze the current response for factual accuracy, logical consistency, adherence to instructions, writing quality, potential improvements, and any biases or unfounded assumptions.
    \item Based on your analysis, revise the response to produce a more accurate, coherent, and polished version that corrects identified issues while maintaining the original intent.
    \item Ensure the final response is clear, accurate, concise, well-structured, and directly addresses the query. It should be logically coherent, contextually relevant, and easy to read. Eliminate all forms of redundancy, such as repetitive information, unnecessary restatements, or duplicated content.
\end{enumerate}
Query: {x}\newline
Response: $y_t$"
\end{quote}

To encourage diversity and robustness, a \textit{best-of-\( n \)} sampling strategy can again be used, functioning as a form of random restart at each iteration. From the refinement prompt, \( n \) candidates are sampled:
\[
\{y_1', \dots, y_n'\} \sim \pi(y \mid x_{\text{refine}}(x, y_t)),
\]
and the candidate with the highest reward is selected:
\[
y^* = \underset{{y' \in \{y_1', \dots, y_n'\}} }{\arg\max} \, r(x, y').
\]
We then compute the reward difference as \( \Delta r = r(x, y^*) - r(x, y_t) \), which determines whether the new candidate is accepted. If \( y^* \) improves upon \( y_t \) (i.e., \( \Delta r \ge 0 \)), it is always accepted. If the reward decreases (\( \Delta r < 0 \)), the new candidate may still be accepted with a probability given by the \textit{Metropolis criterion}:
\[
P(\text{accept}) = 
\begin{cases}
1, & \text{if } \Delta r \geq 0\\
\exp\left(-\dfrac{\Delta r}{T_t}\right), & \text{if } \Delta r < 0,
\end{cases}
\]

where temperature \(T_t\) is updated at each step using a geometric decay schedule:
\[
T_{t+1} = \alpha T_t, \quad \text{with } \alpha \in (0, 1).
\]
This probabilistic acceptance criterion allows non-greedy exploration, enabling the search to escape local optima and increasing the likelihood of discovering globally optimal responses. At high temperatures, ANNETRON explores more widely, frequently accepting worse responses to traverse the response landscape. As the temperature decreases, the algorithm becomes increasingly greedy, primarily accepting responses that yield reward improvements. 

Throughout the optimization process, we maintain a history set \( \mathcal{H} \) of all accepted responses, beginning with the initial candidate and including each accepted refinement:
\[
\mathcal{H} = \{ y_0 \} \cup \{ y^*_1, y^*_2, \dots, y^*_{L} \},
\]
where each \( y^*_t \) denotes a response accepted at iteration \( t \), and \( L \) is the last iteration index.

The process continues for a fixed number of steps or until a termination condition is met (e.g., reward plateau or convergence). The best response observed across all accepted candidates is returned as the final output:
\[
\hat{y} = \arg\max_{y \in \mathcal{H}} \, r(x, y).
\]

\begin{remark}[Analysis of ANNETRON Optimization Dynamics]
ANNETRON is designed to optimize LLM-generated responses under black-box reward functions by combining structured local search with probabilistic exploration. The \emph{\textbf{LLM-as-Refinement-Operator}} edits the current response with context, enabling localized exploration in semantically relevant directions.

Best-of-\(n\) sampling acts as a local randomized reset operator, encouraging steepest-ascent moves by selecting the highest-reward candidate among multiple samples. This increases the chance of discovering better responses within each neighborhood, while still introducing variation across iterations.

The Metropolis acceptance criterion allows ANNETRON to accept worse responses probabilistically, especially at higher temperatures. This helps the algorithm escape local optima and explore a broader region of the response space. As the temperature decays geometrically, the search becomes greedier over time, focusing more on exploiting known high-reward areas.

\end{remark}

\begin{algorithm}[htbp]
\caption{ANNETRON: Simulated Annealing for Response Optimization}
\label{alg:annetron}
\begin{algorithmic}[1]
\STATE Initialize $t \gets 0$
\STATE Sample initial response $y_0 \sim \pi(y \mid x)$
\STATE Evaluate reward $r_0 = r(x, y_0)$
\STATE Set temperature $T_0 > 0$, decay factor $\alpha \in (0, 1)$
\STATE Initialize history buffer $\mathcal{H} \gets \{y_0\}$
\WHILE{termination criterion not met}
    \STATE Construct refinement prompt $x_{\text{refine}} \gets x_{\text{refine}}(x, y_t)$
    \STATE Sample $n$ candidates $\{y_1', \dots, y_n'\} \sim \pi(y \mid x_{\text{refine}})$
    \STATE Select best candidate $y^* = \arg\max_{y' \in \{y_1', \dots, y_n'\}} r(x, y')$
    \STATE Compute reward difference $\Delta r = r(x, y^*) - r(x, y_t)$
    \IF{$\Delta r \geq 0$}
        \STATE Accept: $y_{t+1} \gets y^*$
    \ELSE
        \STATE Accept with probability $p = \exp(-\Delta r / T_t)$
        \IF{\texttt{random()} $< p$}
            \STATE $y_{t+1} \gets y^*$
        \ELSE
            \STATE $y_{t+1} \gets y_t$
        \ENDIF
    \ENDIF
    \STATE Update temperature: $T_{t+1} \gets \alpha T_t$
    \IF{$y_{t+1} \neq y_t$} 
        \STATE Update history: $\mathcal{H} \gets \mathcal{H} \cup \{y_{t+1}\}$
    \ENDIF
    \STATE $t \gets t + 1$
\ENDWHILE
\STATE \textbf{Output:} History buffer $\mathcal{H}$ and best response $\hat{y} = \arg\max_{y \in \mathcal{H}} r(x, y)$
\end{algorithmic}
\end{algorithm}

\subsection{MEMETRON (Memetic Mechanism for Response Optimization)}

To further improve response optimization, we propose \textbf{MEMETRON}, a memetic algorithm that integrates the global exploration capabilities of GENETRON with the local refinement power of ANNETRON. At each generation \( g \), GENETRON produces an intermediate offspring population \( O^{(g)} = \{y_1^*, \dots, y_N^*\} \) via selection, crossover, and mutation implemented implicitly through sampling.

Each offspring \( y^* \in O^{(g)} \) is then refined using ANNETRON, which performs explicit local search to further optimize the candidate with respect to the reward function \( r(x, y) \). After local refinement, MEMETRON proceeds with the remaining GENETRON steps, including updating the memory buffer \( \mathcal{H} \) and applying elitism.

\begin{algorithm}[htbp]
\caption{MEMETRON}
\label{alg:memetron}
\begin{algorithmic}[1]
\STATE $g \gets 0$
\STATE Initialize population $G^{(0)} = \{y_1^{(0)}, \dots, y_N^{(0)}\}$ by sampling from model $\pi(y \mid x)$
\STATE Evaluate rewards and initialize history buffer: $\mathcal{H} \gets G^{(0)}$
\WHILE{termination criterion not met}
    \STATE Apply GENETRON's genetic operators to $G^{(g)}$ to produce offspring $O^{(g)} = \{y_1^*, \dots, y_N^*\}$
    \FORALL{$y^* \in O^{(g)}$}
        \STATE Refine $y^*$ using ANNETRON with respect to reward $r$
    \ENDFOR
    \STATE Update history: $\mathcal{H} \gets \mathcal{H} \cup O^{(g)}$
    \STATE Form next population:
    \[
    G^{(g+1)} = \underset{\substack{S \subseteq \mathcal{H} \\ |S| = N}}{\arg\max} \sum_{y \in S} r(x, y)
    \]
    \STATE $g \gets g + 1$
\ENDWHILE
\STATE \textbf{Output:} $\mathcal{H}$ and $\hat{y} = \arg\max_{y \in \mathcal{H}} r(x, y)$
\end{algorithmic}
\end{algorithm}

\section{Applications}

MEMETRON is broadly applicable to both inference and training workflows. Its flexibility enables deployment in test-time environments where model weights are fixed, as well as in data generation and optimization steps during supervised or reinforcement learning-based training.

\paragraph{Test-Time Optimization}
\label{sec:testtime_compute} MEMETRON is especially effective as a test-time compute mechanism for tasks involving complex reasoning, decision making, or user-specific goals. It enables deliberate exploration of candidate outputs using black-box reward functions, without requiring any model updates or internal access. This is particularly valuable for improving proprietary or large-scale LLMs, where model weights, architecture, or training data are inaccessible. To ensure efficiency, it can be invoked selectively via LLM routers or heuristics that detect ambiguous or high-stakes inputs, rather than being applied to all tasks uniformly. MEMETRON integrates smoothly with existing systems such as reranking, prompt-based strategies, or agent-based planning, and serves as a modular optimization layer for improving generation quality where standard decoding is insufficient.

\paragraph{Training-Time Integration}
\label{sec:training_applications}

The same mechanisms used in MEMETRON can also enhance model training pipelines. In supervised fine-tuning (SFT), responses are typically sampled from a base model using greedy or nucleus decoding. By applying reward-guided search to generate higher-quality completions, MEMETRON improves the training signal without requiring additional human annotations. In RLHF settings, it can replace standard sampling to better explore the reward landscape and construct stronger candidate sets for ranking or preference modeling. For Direct Preference Optimization (DPO), which operates on preference pairs without needing an explicit reward model, MEMETRON can still offer benefits when a reward function is available. Reward model(s) are often used in practice to curate data for DPO. In such cases, it can be used to generate stronger positive and negative examples by searching for high- and low-reward outputs, improving the quality of training pairs beyond simple sampling. In GRPO, where group-wise comparisons drive updates, MEMETRON helps construct wider and informative candidate sets, leading to stronger learning signals across the reward distribution.

\paragraph{Reward Model Analysis} 
Our framework can be used to probe and analyze reward models. By searching the output space to maximize reward signals, MEMETRON is capable of uncovering cases of reward hacking where high-reward responses exploit flaws or misalignments in the reward function.

\paragraph{Other Modalities}

While MEMETRON is designed and evaluated in the context of language models, the underlying principles extend naturally to other generative modalities. For example, models like OpenAI’s Sora, DALL·E, Stable Diffusion, Midjourney, and Flux, often generate multiple candidate images or videos for a given prompt, allowing users to select the most suitable output based on perceptual quality, aesthetic metrics, or downstream task performance. This manual selection process reflects a latent reward signal that could instead be formalized. In such settings, our framework could help in guiding generation toward high-reward outputs without retraining the generative model.

\section{Experiments}\label{experiments}

We evaluate our algorithm on the task of optimizing language model responses using the black-box reward model \texttt{PairRM}, which predicts human preferences between responses. While our method is task-agnostic and applicable to a wide range of reward formulations, we apply it to preference modeling, a practically important and still relatively challenging problem, to demonstrate its flexibility and effectiveness in handling arbitrary black-box reward signals.

\paragraph{Evaluation Dataset}
We use the \texttt{tinyAlpacaEval} dataset, which comprises 100 examples drawn from the full 805‐example \texttt{AlpacaEval 2.0} set. It is designed to offer a quick, low‐cost benchmark for LLMs while preserving the core evaluation properties of the original AlpacaEval 2.0 suite.

\paragraph{Experimental Details}  
We use \texttt{Llama-3.2-3B-Instruct} for all experiments, with fixed sampling hyperparameters: a temperature of 1.5, minimum probability (min-\(p\)) of 0.1, top-\(K = 50\), and a token limit of 4098. For each question in \texttt{tinyAlpacaEval}, we first generate 16 original responses using the above sampling settings. We then apply \texttt{MEMETRON} to evolve these responses over three generations. 

For each generation, we apply \texttt{GENETRON} to evolve the response population. We construct binary tournaments by sampling distinct response pairs without replacement, avoiding duplicate unordered pairs. Winners are selected using the \texttt{PairRM} scoring function. From the pool of winners, we form parent pairs and generate new candidates by sampling from the model conditioned on a fusion prompt \( x_{\text{fuse}}(x, y_i, y_j) \). For each parent pair, we sample three candidates and select the one with the highest \texttt{PairRM} score (best-of-3). This process is repeated for three generations.

Following \texttt{GENETRON}, each selected response is further refined using \texttt{ANNETRON}. In this phase, the goal is to iteratively improve each response while avoiding premature convergence to poor local optima. For each response, we run 7 refinement steps with a patience of 3. At each step, we sample three candidate responses from the language model conditioned on a refinement prompt \( \{y_1', y_2', y_3'\} \sim \pi(y \mid x_{\text{refine}}) \) and select the best out of them, implementing a form of stepwise random restart. Sampling is performed with a temperature schedule that decays geometrically across steps, following \( T_{t+1} = \alpha T_t \) with \( \alpha \in (0, 1) \). To evaluate candidates for preference learning, \texttt{PairRM} is used to compare each sample against the original (pre-annealing) response, which serves as a fixed anchor for all comparisons during refinement. The candidate with the highest preference score relative to the anchor is selected at each step. Anchoring ensures stable and consistent pairwise judgments across steps, preventing drift that can arise when the comparison baseline changes over time.

At the end of the evolution process, we obtain a history buffer \( H \) containing 64 responses, including the original 16 samples generated before evolution.

\paragraph{Statistical Analysis}
For each question, we feed all 64 model-generated responses into PairRM reranking function to get a scalar logit score to each response, reflecting its relative quality within the question’s response set.

To evaluate performance progression, we conduct pairwise statistical comparisons between Generation 4 and each of Generations 1, 2, and 3. We assess normality using the Shapiro–Wilk test. If both groups satisfy the normality assumption ($p > 0.05$), we apply Welch’s \textit{t}-test; otherwise, we use the Mann–Whitney U test. For each comparison, we compute the mean logit score difference, Cohen’s $d$ (standardized effect size), and Cliff’s $\delta$ (ordinal effect size). To control for multiple comparisons across the 100 questions, we apply the Benjamini–Hochberg procedure and report both raw and FDR-adjusted significance levels. See Table~\ref{tab:stats_summary} for a summary of the statistical tools used.

\begin{table}[ht]
\centering
\caption{Statistical tests and effect size metrics used in our analysis.}
\begin{tabularx}{\linewidth}{lX}
\toprule
\textbf{Statistic / Test} & \textbf{Purpose and Description} \\
\midrule
Shapiro–Wilk Test & Tests whether data are normally distributed; used to select between Welch’s and Mann–Whitney tests. \\
Welch’s \textit{t}-test & Compares means between two groups assuming unequal variances (parametric). \\
Mann–Whitney U Test & Compares distributions between two groups (non-parametric alternative to \textit{t}-test). \\
Cohen’s $d$ & Standardized difference in means; measures magnitude of effect (assumes interval scale and normality). \\
Cliff’s $\delta$ & Ordinal effect size; reflects dominance of one group over another without distributional assumptions. \\
Benjamini–Hochberg FDR & Controls the false discovery rate when performing multiple comparisons. \\
\bottomrule
\end{tabularx}
\label{tab:stats_summary}
\end{table}

\paragraph{Results}

\begin{table}[ht]
\small
\centering
\caption{Comparison of Generation 4 with earlier generations across 100 questions. We report average pairwise score differences (logit units), effect sizes, and statistical significance (before and after FDR correction). Welch’s \textit{t}-test or Mann–Whitney U test was selected based on Shapiro–Wilk normality.}
\begin{tabularx}{\linewidth}{lccccX}
\toprule
\parbox[c]{2.5cm}{ \textbf{Compared\\Generations}} & \textbf{Mean Diff ± SD} & \textbf{Welch / MW} & \textbf{Sig. (Raw / FDR)} & \textbf{Cohen’s $d$} & \textbf{Cliff’s $\delta$} \\
\midrule
Gen 1 vs 4 & 11.78 ± 5.07 & 73 / 27 & 93 / 93 & –2.92 & –0.87 \\
Gen 2 vs 4 & 4.60 ± 2.86 & 76 / 24 & 78 / 73 & –1.16 & –0.57 \\
Gen 3 vs 4 & 1.96 ± 1.64 & 81 / 19 & 33 / 15 & –0.55 & –0.31 \\
\bottomrule
\end{tabularx}
\label{tab:group4_comparison}
\end{table}

\begin{figure}[ht]
    \centering

    \begin{subfigure}[t]{0.48\textwidth}
        \centering
        \includegraphics[width=\linewidth]{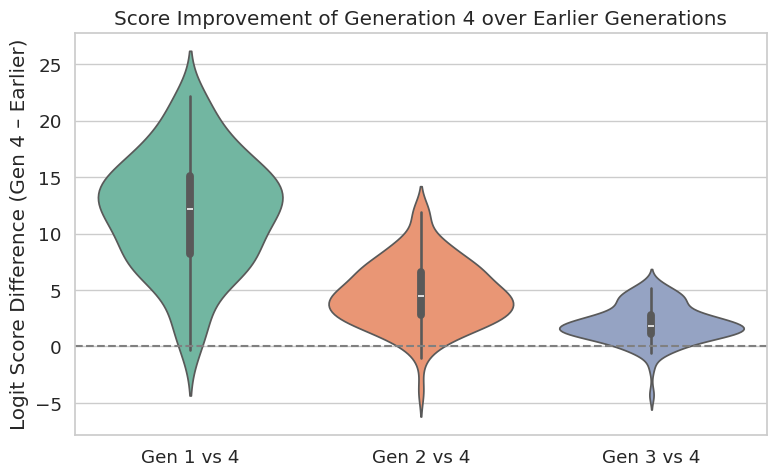}
        \caption{Logit Score Differences (Gen 4 – Earlier)}
        \label{fig:violin_plot}
    \end{subfigure}
    \hfill
    \begin{subfigure}[t]{0.48\textwidth}
        \centering
        \includegraphics[width=\linewidth]{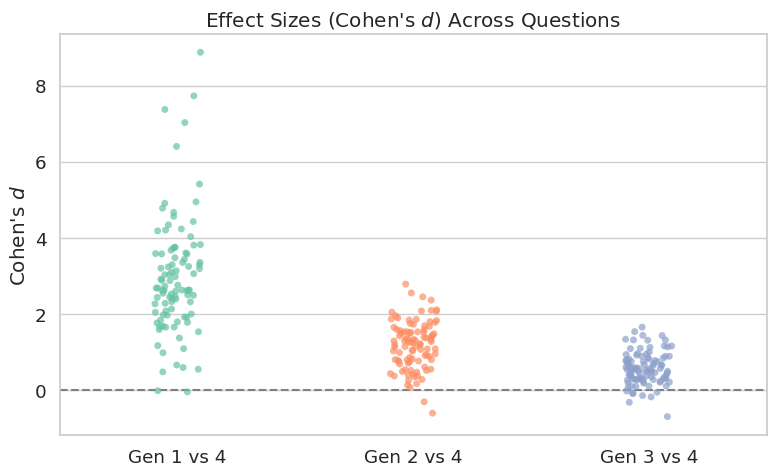}
        \caption{Cohen’s $d$ Effect Sizes}
        \label{fig:cohen_d_plot}
    \end{subfigure}

    \vspace{1em}

    \begin{subfigure}[t]{0.48\textwidth}
        \centering
        \includegraphics[width=\linewidth]{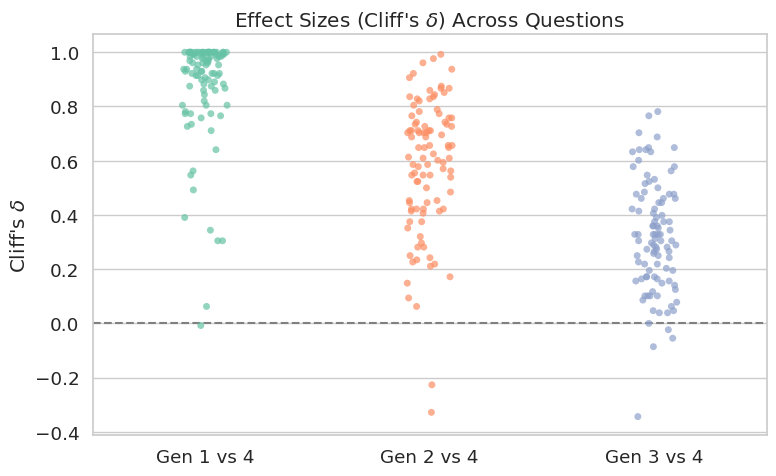}
        \caption{Cliff’s $\delta$ Ordinal Effect Sizes}
        \label{fig:cliffs_delta_plot}
    \end{subfigure}

    \caption{
    Performance comparisons between Generation 4 and earlier generations across 100 questions.
    \textbf{(a)} Violin plot of score differences shows that Gen 4 consistently improves upon earlier generations, especially Gen 1.
    \textbf{(b)} Cohen’s $d$ indicates large standardized effect sizes for most questions, particularly for Gen 1 and Gen 2.
    \textbf{(c)} Cliff’s $\delta$ confirms ordinal dominance of Gen 4, showing strong pairwise superiority over Generations 1 and 2, with reduced dominance by Generation 3.
    }
    \label{fig:comparison_analysis}
\end{figure}

We analyze model performance progression by comparing the final generation (Generation 4) to each earlier generation across 100 questions. Table~\ref{tab:group4_comparison} summarizes the results.

The largest improvement is observed between Generation 1 and Generation 4, with an average score increase of \textbf{11.78 ± 5.07} logits. A total of \textbf{93\% of questions} remain statistically significant after FDR correction. The effect sizes are large and consistent across questions (\textbf{Cohen’s $d$ = –2.92}, \textbf{Cliff’s $\delta$ = –0.87}), indicating substantial gains in relative response quality. Generation 2 also shows a strong improvement (\textbf{4.60 ± 2.86} logits, \textbf{73\% significant}, \textbf{Cohen’s $d$ = –1.16}, \textbf{Cliff’s $\delta$ = –0.57}). In contrast, the difference between Generation 3 and Generation 4 is smaller (\textbf{1.96 ± 1.64} logits), and only \textbf{15\% of questions} remain significant after correction (\textbf{Cohen’s $d$ = –0.55}, \textbf{Cliff’s $\delta$ = –0.31}).

The results show that response quality improves most sharply in the early stages, with diminishing returns in later generations. Generation 4 outperforms Generation 1 with large effect sizes and near-universal statistical significance. While improvements over Generation 2 remain notable, they are smaller in magnitude. By Generation 3, gains are marginal and less consistent, with significance dropping to 15\%. This pattern suggests that most of the benefit is realized early.

Figure~\ref{fig:comparison_analysis} illustrates the overall progression of response quality across generations. The plots confirm that the most substantial improvements occur early in the sequence, with Generation 4 outperforming earlier generations, particularly Generation 1, with diminishing gains by Generation 3.

The experiments show that MEMETRON significantly outperforms zero-shot decoding and best-of-16 sampling by more effectively leveraging the same 16 generated responses to explore and refine the output space.

\section{Conclusions}\label{conclusions}

We introduced MEMETRON, a framework for metaheuristic test-time optimization of LLM responses. By formulating inference as a discrete search problem over candidate completions, MEMETRON enables reward-guided exploration without requiring changes to model parameters. Its hybrid algorithms, GENETRON, ANNETRON, and MEMETRON, combine the generative strengths of LLMs with classical search strategies, supporting broad applicability across domains and tasks.

In addition to test-time refinement, the same principles can enhance training workflows such as SFT, RLHF, DPO, and GRPO by improving the quality of training targets and optimization signals. The framework supports a wide range of modalities and reward types, and can also serve as a tool for probing reward models to identify misalignments or reward hacking behaviors. As language models are increasingly used in complex and adaptive applications, MEMETRON offers a practical foundation for aligning their outputs with diverse user-defined goals.

\section{Limitations and Future Work}

Our approach assumes access to a reasonably capable base language model. If the underlying LLM lacks fluency, coherence, or task competence, the benefits of response-level search may be limited. Additionally, MEMETRON relies on user-defined reward functions. While flexible, this assumes that users can construct meaningful and reliable reward models, which may be challenging in subjective or open-ended domains.

Future work includes extending MEMETRON to handle multi-objective optimization, allowing the framework to balance trade-offs such as factuality, style, efficiency, and safety simultaneously. We also aim to explore integration with adaptive routing and scheduling systems, enabling more intelligent allocation of compute during deployment.

\newpage
\bibliography{refs_ranking}
\bibliographystyle{unsrt}
\newpage

\appendix
\end{sloppypar}
\end{document}